\documentclass{article}

\usepackage{graphicx}
\usepackage{booktabs}
\usepackage{multirow}
\usepackage{arydshln}
\usepackage{tabularray}
\usepackage{booktabs}
\usepackage{longtable}
\usepackage{array}
\usepackage{listings}
\usepackage{float}
\usepackage{enumitem}   

\newcolumntype{L}[1]{>{\raggedright\arraybackslash}p{#1}}

\usepackage{siunitx}   

\usepackage[final]{neurips_2024}

\usepackage{caption} 
\usepackage[T1]{fontenc}    
\usepackage{hyperref}       
\usepackage{url}            
\usepackage{booktabs}       
\usepackage{amsmath}        
\usepackage{amsfonts}       
\usepackage{nicefrac}       
\usepackage{microtype}      
\usepackage{xcolor}         

\usepackage{booktabs}      
\usepackage{pifont}        
\usepackage[utf8]{inputenc} 
\usepackage{tikz}
\usepackage{xspace}
\usetikzlibrary{arrows.meta,positioning,shapes.geometric,calc}

\title{Signals: Trajectory Sampling and Triage for Agentic Interactions}

\author{%
  Shuguang Chen \hspace{2em} Adil Hafeez \hspace{2em} Salman Paracha \\ [0.5ex]
  DigitalOcean Holdings, Inc. \\ [0.5ex]
  \texttt{\{schen, ahafeez, salmanparacha\}@digitalocean.com} \\
}

\date{}

\begin{document}
\maketitle


\begin{abstract}

Agentic applications based on large language models increasingly rely on multi-step interaction loops involving planning, action execution, and environment feedback. While such systems are now deployed at scale, improving them post-deployment remains challenging. Agent trajectories are voluminous and non-deterministic, and reviewing each one, whether through human review or auxiliary LLMs, is slow and cost-prohibitive. We propose a lightweight, signal-based framework for triaging agentic interaction trajectories. Our approach computes cheap, broadly applicable signals from live interactions and attaches them as structured attributes for trajectory triage, identifying interactions likely to be informative without affecting online agent behavior. We organize signals into a coarse-grained taxonomy spanning interaction (misalignment, stagnation, disengagement, satisfaction), execution (failure, loop), and environment (exhaustion), designed for computation without model calls. In a controlled annotation study on $\tau$-bench, a widely used benchmark for tool-augmented agent evaluation, we show that signal-based sampling achieves an 82\% informativeness rate compared to 74\% for heuristic filtering and 54\% for random sampling, with a 1.52× efficiency gain per informative trajectory. The advantage is robust across reward strata and task domains, confirming that signals provide genuine per-trajectory informativeness gains rather than merely oversampling obvious failures. These results show that lightweight signals can serve as practical sampling infrastructure for agentic systems, and suggest a path toward preference data construction and post-deployment optimization.
\end{abstract}

\section{Introduction}
\label{sec:introduction}
Agentic applications combine LLM reasoning with tool execution across multi-step interactions \cite{schick2023toolformer, yao2022react}. As these systems move into real-world deployment, improving them requires turning observed behavior into optimization signal. Deployed agents generate detailed behavioral data at scale, spanning reasoning steps, tool use, execution outcomes, and user responses. Meanwhile, preference-based training methods such as RLHF \cite{ouyang2022training} and DPO \cite{rafailov2023direct} offer well-studied mechanisms for improving LLM behavior given appropriately constructed preference data. Yet a gap persists between the behavioral data agents produce and the preference-learning methods that could consume it. Production systems capture rich trajectories but provide no mechanism for translating them into training signal. Preference-learning pipelines require curated comparison data but have no systematic way to source it from production. The result is that improving deployed agents remains predominantly manual. Developers inspect trajectories by hand, hypothesize failure modes, and iterate on prompts or tool definitions without a structured pipeline connecting the two.

This disconnect persists for several reasons. Offline evaluation, while valuable for testing known scenarios, relies on curated benchmarks that miss the long tail of real-world usage \cite{yao2024tau, pan2026measuring}. Manual review does not scale. Agent trajectories are voluminous and non-deterministic, and there is no reliable indicator to detect when an agent loops unproductively or a user grows frustrated. A natural alternative is to evaluate every trajectory with an auxiliary LLM, as LLM-as-a-judge approaches \cite{zheng2023judging} have shown over 80\% agreement with human preferences on structured tasks, but applying such evaluation to every trajectory is cost-prohibitive at scale. Prior work on dialogue quality has proposed automatic quality indicators derived from conversation features \cite{Higgins2024ACQI, Schmitt2015IQ}, but these approaches make assumptions that do not hold for agentic systems. First, they treat conversation as the whole picture, whereas agents interleave a discourse layer (user intent, clarifications, frustration) with an execution layer (tool calls, API responses, state changes). An agent can maintain fluent and friendly conversation while catastrophically failing at execution. Second, they assume signals should produce quality scores or prescribe fixes, but quality judgments are context-dependent. A terse response may be ideal for an expert user but frustrating for a novice, and embedding such judgments into the system risks encoding assumptions that do not generalize across domains.

We propose to bridge this disconnect using lightweight trajectory signals composed into a triage function. Signals are descriptive markers of recurring behavioral patterns, spanning interaction (misalignment, stagnation, disengagement, satisfaction), execution (failure, loop), and environment (exhaustion), that can be computed without model calls and attached to trajectories as structured metadata. Interaction and execution signals are learning-oriented, suitable for constructing preference data, while environment signals support system-level diagnosis without serving as training supervision. Crucially, signals are not quality scores. They identify trajectories that are likely to be informative for downstream analysis, surfacing both failures and exemplars, without asserting correctness or prescribing remediation. This design draws on a long tradition in information retrieval, where implicit behavioral signals such as query reformulation, dwell time, and session abandonment have served as proxies for user satisfaction without requiring explicit feedback \cite{DBLP:journals/tois/FoxKMDW05, joachims2002optimizing}. We adapt this idea to the agentic setting, where trajectories contain not only natural language but also tool calls, execution outcomes, and environment feedback.


Concretely, this paper makes three contributions:

\begin{enumerate}[topsep=0ex,itemsep=0ex,partopsep=-0.5ex,parsep=0.5ex]
    \item \textbf{A signal taxonomy for agentic systems} spanning interaction (misalignment, stagnation, disengagement, satisfaction), execution (failure, loop), and environment (exhaustion) signals, designed for computation without model calls.
    \item \textbf{A sampling framework that avoids quality scoring}. We introduce an aggregation scheme and parallel sampling streams for failures and exemplars that prioritize trajectories for human review without computing quality metrics or prescribing actions.
    \item \textbf{Empirical validation on $\boldsymbol{\tau}$-bench}, a widely used benchmark for tool-augmented agent interactions, showing that signal-based sampling achieves an 82\% informativeness rate (vs. 54\% random, 74\% heuristic) with a 1.52× efficiency gain, and that this advantage is robust across reward strata and task domains.
\end{enumerate}

\section{Related Work}
\paragraph{Agentic Systems and Post-Deployment Improvement.}
Modern agentic systems build on LLMs that interleave reasoning with action execution in multi-step loops. ReAct \cite{yao2022react} generates interleaved reasoning trajectories and task-specific actions, while Toolformer \cite{schick2023toolformer} trains LLMs to autonomously invoke external tools via simple APIs. Improving such agents after deployment remains an open challenge. Reflexion \cite{shinn2023reflexion} reinforces language agents through verbal self-reflection stored in episodic memory. Self-Refine \cite{madaan2023self} iteratively improves LLM outputs by having the same model generate feedback and refine itself. ExpeL \cite{zhao2024expel} autonomously gathers experiences and extracts natural-language insights from training tasks for use at inference. Voyager \cite{wang2024voyager} builds a skill library of executable code through iterative prompting with environment feedback. These methods operate \emph{online} within episodes or across training tasks in simulated environments and require ground-truth rewards, task retries, or environment resets, conditions often unavailable in real-world deployments. Our work instead processes completed trajectories from deployed systems offline without modifying the agent's online behavior.

\paragraph{Preference Learning and Data Construction.}
Preference-based training is the dominant paradigm for aligning LLMs. RLHF \cite{ouyang2022training} fine-tunes language models using labeler demonstrations and output rankings. Direct Preference Optimization (DPO) \cite{rafailov2023direct} simplifies this by extracting the optimal policy via a classification loss without explicit reward modeling. While these training methods are well-studied, sourcing preference pairs remains a bottleneck for agentic settings. LLM-as-a-judge approaches \cite{zheng2023judging} show that strong LLM judges can match human preferences with over 80\% agreement, but applying them to every trajectory from a deployed system is cost-prohibitive. Agent-as-a-Judge \cite{zhuge2025agentasajudge} extends this paradigm to agentic systems by incorporating intermediate feedback across the task-solving process, but still requires model calls for each evaluated trajectory. Process reward models \cite{lightman2024lets} offer step-level supervision but require expensive annotation or verifiable rewards. Our work addresses the data construction side: lightweight trajectory signals triage which trajectories merit review, and we propose a path toward transforming selected trajectories into preference pairs via counterfactual continuations.

\paragraph{Automatic Quality Signals from Behavioral Data.} 
When explicit user feedback or ground-truth rewards are unavailable, systems must rely on signals derived from the trajectory itself. In dialogue systems, the Dialogue Breakdown Detection Challenge \cite{higashinaka-etal-2016-dialogue} organized shared tasks to identify inappropriate system utterances using features such as dialogue-act types and coherence measures. In information retrieval, implicit signals such as query reformulation and session abandonment have long served as proxies for user satisfaction. In the agentic setting, Watson \cite{DBLP:conf/kbse/RombautMVLH25} introduces cognitive observability for LLM-powered agents, recovering implicit reasoning traces via a surrogate model to support diagnosis and targeted correction, but requires model calls per trajectory. Our behavioral signals, including repetition, repair, and frustration patterns, adapt implicit-signal ideas from dialogue and retrieval to the agentic setting, where trajectories contain not only natural language but also tool calls, execution outcomes, and environment feedback. Our execution-level signals, such as tool-call failures and failure patterns, are straightforward to extract from structured trajectory records but remain underutilized as training signals. While prior work has separately studied agent architectures, self-improvement, preference optimization, and observability, less attention has been paid to the interface between observability and optimization. Our work targets this gap: we compose lightweight, model-free trajectory signals into a triage function over heterogeneous agent trajectories, selecting developer-informative trajectories that can serve as the basis for preference data construction.

\section{Signal Taxonomy}
\label{sec:signal_taxonomy}

We propose a coarse-grained taxonomy of trajectory signals and a lightweight detection procedure that instantiates these signals from observable trajectories. We define signals as descriptive markers of recurring behavioral patterns that make a trajectory potentially informative for downstream analysis and preference-data construction. Importantly, they are not intended to measure correctness, reward, or quality. We organize trajectory signals along two orthogonal axes: the data layer from which they are derived and their downstream utility. The first axis distinguishes between signals derived from the discourse layer of a trajectory (user–assistant natural language) and those derived from the execution layer (e.g., tool calls, runtime events). The second axis distinguishes between signals that are useful for learning (i.e., for constructing preference data and improving agent policies) and signals that are primarily useful for diagnosis and system observability. This yields three top-level groups of signals: 1) \textbf{Interaction Signals} (learning-oriented), 2) \textbf{Execution Signals} (learning-oriented), and 3) \textbf{Environment Signals} (diagnosis-oriented). The categorization reflects differences in how signals are computed and in their intended use rather than a conceptual separation of agent behavior; a single trajectory may simultaneously exhibit patterns from multiple groups. The taxonomy is intentionally coarse-grained: we collapse many narrow heuristics into a small set of recurring patterns to generalize across implementations and support scalable triage.

\subsection{Interaction Signals}
\paragraph{Definitions.} Interaction signals are computed from user–assistant natural language and capture how the interaction unfolds at the discourse level. They reflect user-facing behavior and cooperative dynamics, without making claims about internal agent state or semantic correctness. These signals are suitable for preference learning because they expose success and failure modes that are directly legible to users. 

We group interaction signals into four recurring discourse-level patterns:
\begin{itemize}[topsep=0ex,itemsep=0ex,partopsep=-0.5ex,parsep=0.5ex]
    \item \textbf{Misalignment.} Misalignment signals capture semantic or intent mismatch between the user and the agent, such as rephrasing, corrections, clarifications, and restated constraints. Importantly, these signals do not assert that either party is “wrong”; they only indicate that shared understanding has not yet been established.

    \item \textbf{Stagnation.} Stagnation signals capture cases where the discourse continues but fails to make visible progress. This includes near-duplicate assistant responses, circular explanations, repeated scaffolding, and other forms of linguistic degeneration. Unlike execution-level loops, stagnation is defined in terms of discourse dynamics rather than control flow.

    \item \textbf{Disengagement.} Disengagement signals mark the withdrawal of cooperative intent from the interaction. These include explicit requests to exit the agent flow (e.g., “talk to a human”), strong negative stances, and abandonment markers when session boundaries are observable. Disengagement differs from misalignment and stagnation in that it represents a terminal or near-terminal state.

    \item \textbf{Satisfaction.} Satisfaction signals indicate successful convergence and completion of the interaction. These include expressions of gratitude, success confirmations (e.g., “that worked”), and closing utterances. We use these signals to sample exemplar trajectories rather than to assign quality scores.
\end{itemize}

\paragraph{Detection.} Interaction signals are detected using lightweight normalization and interpretable, typo-tolerant matching over user turns. Misalignment, disengagement, and satisfaction are primarily triggered by phrase-level cues, with additional local similarity checks across nearby turns to capture rephrasing even when explicit markers are absent. Stagnation is detected using simple discourse heuristics that summarize repetition and inefficiency (e.g., near-duplicate phrasing within a speaker role and prolonged interactions relative to a baseline). The overall design emphasizes robustness to surface variation while keeping triggers traceable to specific message spans for triage.

\subsection{Execution Signals}
\paragraph{Definitions.} Execution signals are derived from structured runtime events emitted by the agent’s internal control loop. These may include reasoning steps, action selections, tool or web invocations, memory operations, or other agentic actions. Unlike interaction signals, execution signals are modality-independent and typically deterministic. We isolate execution signals as a separate class because they reflect agent decision-making behavior rather than external system conditions. 

We group execution signals into two recurring behavioral patterns:
\begin{itemize}[topsep=0ex,itemsep=0ex,partopsep=-0.5ex,parsep=0.5ex]
    \item \textbf{Failure.} Failure signals capture action attempts that do not yield a usable or task-advancing outcome (e.g., empty results, no-op actions, inappropriate action choices), without attributing blame to the agent or the environment. These signals are learning-relevant because they shape the agent's subsequent behavior.

    \item \textbf{Loop.} Loop signals capture repetitive execution patterns in which the agent remains active but does not make progress. These include retries, oscillations between strategies or action types, and progressive parameter drift. These patterns are treated uniformly as manifestations of non-progressing control flow.
\end{itemize}

\paragraph{Detection.} Execution failures are detected by classifying non-advancing tool outcomes from structured observations and associating each outcome with its triggering invocation to retain relevant context (e.g., tool identity and arguments). Execution loops are detected via sequence analysis over invocation streams, using simple pattern rules that identify repeated calls with identical inputs, repeated calls with systematically varying inputs, and repeated multi-tool cycles. This separation allows failures to capture localized breakdowns while loops capture sustained non-progressing control flow.

\subsection{Environment Signals}
\paragraph{Definitions.} Environment signals capture failures and constraints arising from the surrounding system rather than the agent’s internal policy or reasoning. These include infrastructure, API, and resource-boundary conditions. We isolate these signals because, while they are essential for observability and diagnosis, they are not suitable as training supervision. They do not reflect the quality of the agent’s decisions and can introduce spurious correlations if used for learning. Note that if the event is primarily explained by system constraints or service health (quota, outage, context cap), we classify it as Environment; otherwise, we classify it as Execution. 

We group environment signals into a single high-level pattern:
\begin{itemize}[topsep=0ex,itemsep=0ex,partopsep=-0.5ex,parsep=0.5ex]
    \item \textbf{Exhaustion.} Exhaustion signals capture boundary and infrastructure conditions, such as context overflows, rate limits, API failures, and malformed external responses, that terminate or degrade behavior independently of agent competence. They are used for diagnosis rather than learning.
\end{itemize}

\paragraph{Detection.} Exhaustion is detected from tool observations by identifying external failure and resource-limit indicators in system outputs. The detector produces trajectory-localized instances that support diagnosis and system-level triage, and it distinguishes environment-driven constraints from execution-driven issues by attributing events to external service conditions and resource boundaries when those indicators are dominant.

\section{Experiment}
\label{sec:experiment}
We evaluate the proposed signal framework as a data selection mechanism that can sit upstream of preference construction and training. Rather than assessing signals as classifiers or quality scorers, we ask whether they can serve as practical sampling infrastructure: identifying which trajectories merit human review, without requiring semantic understanding or explicit reward modeling. Concretely, we aim to validate the claim that signal sampling surfaces a higher fraction of developer-informative trajectories than baseline strategies at a fixed annotation budget, revealing meaningful success or failure modes that would otherwise be diluted or missed entirely.


\subsection{Experiment Setup}

\paragraph{Trajectory Pool. }  We use $\tau$-bench \cite{yao2024tau} as our testbed, a benchmark that emulates multi-turn conversations between a simulated user and a tool-equipped agent across two domains (i.e., airline and retail). Its trajectories contain both discourse-level interaction (user–agent dialogue) and structured execution events (tool calls, API responses, database mutations), exercising all signal categories in the proposed framework. We construct the trajectory pool from $\tau$-bench's publicly available historical trajectories, generated by multiple agent configurations (varying model backbones and prompting strategies) across all benchmark tasks. The resulting pool naturally includes both successes and failures. We denote the total pool size as $N$ and draw fixed-size samples of $n  = 100$ trajectories per method. We note that $\tau$-bench uses LLM-simulated users rather than real ones; certain interaction signals, particularly disengagement and satisfaction, may therefore be under-represented relative to real-world traffic.

\paragraph{Sampling Strategies.} We compare three sampling methods, each drawing 100 trajectories:

\begin{itemize}[topsep=0ex,itemsep=0ex,partopsep=-0.5ex,parsep=0.5ex]
    \item \textbf{Random}: Uniform sampling from the full trajectory pool, serving as the unbiased baseline.
    \item \textbf{Heuristic}: Trajectories containing at least 10 user messages, the most natural first-pass filter a practitioner might apply. This threshold captures the intuition that longer conversations are more complex or more likely to contain failures. However, conversation length is a surface correlate of difficulty, not a direct indicator of informativeness.
    \item \textbf{Signal}: Trajectories selected by the combined interaction and execution signals. Environment signals are excluded, consistent with their diagnosis-only role. This method uses the full set of interaction signals (misalignment, stagnation, disengagement, satisfaction) and execution signals (failure, loop), aggregated into a composite triage score that prioritizes trajectories exhibiting one or more signal activations.
\end{itemize}

All three sampling strategies draw the same number of trajectories ($n = 100$), ensuring that any difference in annotation yield is attributable to the sampling strategy rather than annotation volume.

\paragraph{Annotation Protocol.} Three expert annotators, each familiar with agentic systems and tool-use patterns, independently label all 300 trajectories. Trajectories from all conditions are shuffled into a single queue and annotators are blinded to which sampling strategy produced each trajectory. For each trajectory, annotators answer two labeling questions: 1) \textbf{Developer-informative?} (YES / NO). A trajectory is labeled YES if it contains enough concrete evidence for a developer to form at least one plausible hypothesis about how to improve the agent's behavior, including both failures and strong successes; and 2) \textbf{Main reason} (single-select). One of six categories is labeled for each trajectory: action/tool-use behavior issue, conversation issue, external system issue, success exemplar, none/unclear, or other. Priority rules resolve multi-issue trajectories: external system issues are selected only when an external dependency failure is the dominant driver; action/tool-use issues take priority over conversation issues when execution is the key failure mode. An optional free-text note captures a one-sentence explanation.

\paragraph{Evaluation Metric.} The primary metric is the \textit{informativeness rate}: the fraction of sampled trajectories judged developer-informative by majority vote ($\geq$2 of 3 annotators). Higher rates indicate that a sampling strategy surfaces more useful trajectories per unit of annotation effort.

\subsection{Results and Analysis}

\paragraph{Inter-annotator Agreement.} Before reporting outcomes, we assess annotation reliability. Individual annotator YES-rates range from 0.57 to 0.74, yielding a prevalence index of 0.34 and a bias index of 0.17. Both factors are known to deflate $\kappa$-family statistics \cite{Feinstein1990HighAB} even when raw agreement is adequate; we therefore rely on Gwet's AC1 \cite{Gwet2008ComputingIR}, which corrects for these effects, as the primary agreement coefficient. For the binary \emph{developer-informative} question, three-rater Gwet's AC1 is 0.477, indicating moderate agreement. Because this question requires a binary cut on a graded, subjective property, disagreement concentrates on borderline trajectories where evidence is present but ambiguously actionable; the moderate value thus reflects threshold disagreement rather than a lack of shared understanding. This interpretation is confirmed by the \emph{main-reason} category: conditional on trajectories where all three annotators agree that a trajectory is informative ($N$ = 130), Fleiss' $\kappa$ = 0.662 and Gwet's AC1 = 0.829, indicating that annotators reliably identify the same underlying issue once they agree that a trajectory is worth examining. We use majority-vote aggregation for all subsequent analyses.

\paragraph{Main Results.} Table~\ref{tab:main_result} reports the informativeness rate for each of the three sampling strategies, each evaluated on a fixed budget of 100 trajectories. The informativeness rate is the percentage of sampled trajectories that annotators judged to contain actionable information for developers. All confidence intervals are Clopper--Pearson 95\% intervals, which guarantee exact coverage for binary outcomes, and all pairwise comparisons use Fisher's exact test (two-sided) to ensure validity without relying on asymptotic approximations.

Signal sampling achieves the highest overall informativeness rate
(82.0\%, 95\% CI [.73,\,.89]), meaning that roughly four out of every five trajectories it selects provide useful diagnostic or behavioral evidence. By comparison, heuristic sampling reaches 74.0\% and random sampling only 54.0\%. The difference between signal and random sampling is highly significant ($p < 0.001$), confirming that signal sampling surfaces substantially more developer-informative trajectories at a fixed annotation budget. The difference between signal and heuristic sampling does not reach
significance ($p = 0.232$); however, as we show next, the two
strategies differ sharply in \emph{how} they achieve their rates
once we stratify by task reward.

\begin{table}[t]
\centering
\caption{Informativeness rate by sampling strategy (majority-vote).}
\label{tab:main_result}
\vspace{3pt}
\renewcommand{\arraystretch}{1.2}
\resizebox{0.9\linewidth}{!}{
    \begin{tabular}{l lll lll lll}
        \toprule
        & \multicolumn{3}{c}{\textbf{Overall}} & \multicolumn{3}{c}{\textbf{Failed (reward\,=\,0)}} & \multicolumn{3}{c}{\textbf{Successful (reward\,=\,1)}} \\
        \cmidrule(lr){2-4} \cmidrule(lr){5-7} \cmidrule(lr){8-10}
        Strategy & $N$ & Rate & 95\% CI & $N$ & Rate & 95\% CI & $N$ & Rate & 95\% CI \\
        \midrule
        Random    & 100 & 54.0\% & {[}.44,\;.64{]} & 37 & 75.7\% & {[}.59,\;.88{]} & 63 & 41.3\% & {[}.29,\;.54{]} \\
        Heuristic & 100 & 74.0\%$^{*}$ & {[}.64,\;.82{]} & 70 & 84.3\% & {[}.74,\;.92{]} & 30 & 50.0\% & {[}.31,\;.69{]} \\
        Signal    & 100 & \textbf{82.0\%}$^{*}$ & {[}.73,\;.89{]} & 52 & \textbf{96.2\%}$^{*\dagger}$ & {[}.87,\;1.0{]} & 48 & \textbf{66.7\%}$^{*}$ & {[}.52,\;.80{]} \\
        \bottomrule
    \end{tabular}%
}
\vspace{3pt}

{\small  CI: Clopper–Pearson 95\% interval. Fisher's exact test, two-sided. \\ $^{*}$\,Significantly higher than Random ($p < 0.05$).\quad $^{\dagger}$\,Significantly higher than Heuristic ($p < 0.05$).}

\end{table}

\paragraph{Reward-stratified Analysis.} Each trajectory in $\tau$-bench carries a binary reward: a trajectory is considered successful only if the final database state exactly matches the ground-truth outcome and the agent's response includes all required information. Stratifying by this reward exposes a key compositional difference among the three samplers. As shown in Table \ref{tab:main_result}, heuristic sampling selects predominantly failed trajectories (70\% with reward\,=\,0), while random sampling reflects the pool's base rate (37\% failed). Signal sampling draws a more balanced mix (52\% failed). This compositional difference has important implications.

Among \textit{failed trajectories}, all strategies achieve high informativeness rates (75.7\% -- 96.2\%), with signal sampling reaching 96.2\%. The practical gap is more pronounced among \textit{successful trajectories}, where the three strategies diverge most: signal sampling identifies informative patterns in 66.7\% of \textit{successful trajectories}, compared to 50.0\% for heuristic sampling and only 41.3\% for random sampling. These are the subtle behavioral issues, e.g., policy violations and inefficient tool use, that do not prevent task completion but still matter for improvement.

To isolate each strategy's ability to find informative trajectories from its tendency to over-sample failures, we perform a counterfactual standardization, re-weighting each strategy's stratum-specific rates to the reward distribution of random sampling (63\% success, 37\% fail). Under this adjustment, signal sampling achieves a standardized rate of 77.6\%, compared to 62.7\% for heuristic sampling and 54.0\% for random sampling. The advantage of heuristic sampling over random sampling drops by 11.3 percentage points once its failure-heavy composition is removed, while the advantage of signal sampling is more robust (only a 4.4-point reduction). This confirms that signal sampling provides genuine per-trajectory informativeness gains rather than merely oversampling failed trajectories.

\paragraph{Annotation Efficiency.} The preceding results translate directly into practical savings. At a fixed budget of 100 annotations, signal sampling yields 82 informative trajectories, compared with 74 for heuristic sampling and 54 for random sampling. Equivalently, each informative trajectory costs 1.22 labels under signal sampling, versus 1.35 for heuristic sampling and 1.85 for random sampling, corresponding to a 1.52× efficiency gain over the unbiased baseline. Moreover, as the reward-stratified analysis shows, this gain is not merely an artifact of oversampling obvious failures: signal sampling maintains higher informativeness rates within both the failed and successful strata, meaning the efficiency advantage persists even when the composition of the sample is held constant.

\paragraph{Category Distribution.} Table \ref{tab:category_distribution} presents the distribution of annotated reason categories among developer-informative trajectories. Among informative trajectories, the distribution of annotated reasons is stable across all three strategies: action/tool-use behavior issues account for 57--60\% and conversation issues for 38--43\%, with small numbers of success exemplars. This consistency suggests that the signal framework does not bias the type of issue surfaced and instead simply surfaces more of them.

\paragraph{Domain Robustness.} We additionally examine whether the advantage of signal sampling holds across different domains in $\tau$-bench. In the airline domain, all strategies achieve high informativeness rates (86\% -- 96\%), leaving limited room for differentiation. The retail domain, which features more complex multi-step tasks and a lower base informativeness rate, reveals the clearest separation: signal sampling achieves 78\% informativeness rate compared to 66\% for heuristic sampling and 35\% for random sampling. Signal sampling thus provides the largest marginal value precisely where trajectories are most heterogeneous and uninformative trajectories are most prevalent.

\begin{table}[t]
\centering
\caption{Distribution of main reason among developer-informative traces (majority vote).}
\label{tab:category_distribution}
\vspace{3pt}
\renewcommand{\arraystretch}{1.2}
\resizebox{0.75\linewidth}{!}{
    \begin{tabular}{lcccc}
        \toprule
        Strategy ($N_{\text{informative}}$) & Action / Tool-use & Conversation & Success Exemplar \\
        \midrule
        Random (54)    & 31 (57.4\%) & 23 (42.6\%) & 0 (0.0\%) \\
        Heuristic (74) & 43 (58.1\%) & 28 (37.8\%) & 3 (4.1\%) \\
        Signal (82)    & 49 (59.8\%) & 31 (37.8\%) & 2 (2.4\%) \\
        \bottomrule
    \end{tabular}
}
\end{table}

\section{Limitations}
Our experiments are conducted on $\tau$-bench, which spans two domains (airline and retail) and uses LLM-simulated users. While these domains exercise all signal categories in the taxonomy, whether the observed advantages generalize to a broader range of domains and to real user populations remains an open question. In particular, simulated users may under-represent the variability of real disengagement and satisfaction patterns, as noted in Section \ref{sec:experiment}.

Additionally, the signal taxonomy is intentionally coarse-grained and behavioral. It captures recurring discourse and execution patterns but does not assess semantic correctness or domain-specific policy violations. Trajectories that are fluent and behaviorally unremarkable yet factually wrong will not be surfaced by the current framework, suggesting that signals are best used alongside complementary mechanisms such as domain-specific validators or outcome verification.

Finally, our signal detectors rely on deterministic rules and lexical heuristics. Model-based detectors could offer improved recall, particularly for subtle misalignment or implicit frustration patterns that lack explicit lexical markers, though at the cost of the lightweight computation that makes always-on deployment feasible. Exploring hybrid architectures that combine rule-based signals with selective model-based detection is a promising direction.

\section{Conclusion}
We presented a lightweight, signal-based framework for triaging agentic interaction trajectories. The framework contributes: 1) a coarse-grained signal taxonomy spanning interaction, execution, and environment patterns; 2) a sampling mechanism that prioritizes trajectories for human review without computing quality scores; and 3) empirical validation on $\tau$-bench showing that signal-based sampling achieves an 82\% informativeness rate with a 1.52× efficiency gain over random sampling. The advantage is robust across reward strata and task domains, confirming that signals provide genuine per-trajectory informativeness gains rather than merely oversampling obvious failures.

The framework is not only effective, but also practical for real-world deployment. Since all signals are computed through deterministic rules rather than model calls, the method incurs negligible overhead and scales easily to large collections of interaction traces. This combination of utility and efficiency makes signal-based sampling a compelling first stage in a broader preference data construction pipeline: the selected trajectories, including both failures and successful exemplars, can be paired with counterfactual continuations to produce supervision for preference-based optimization. We leave the design and evaluation of this end-to-end pipeline to future work.

\bibliographystyle{plain}
\bibliography{signals}
\newpage
\appendix




\end{document}